\documentclass{article}

\usepackage{microtype}
\usepackage{graphicx}
\usepackage{subcaption}
\usepackage{booktabs} 
\usepackage{multirow}
\usepackage{graphicx}
\usepackage{arydshln}

\usepackage{hyperref}

\usepackage[preprint]{icml2026}

\usepackage{amsmath}
\usepackage{amssymb}
\usepackage{mathtools}
\usepackage{amsthm}
\usepackage{cuted}

\usepackage[capitalize,noabbrev]{cleveref}

\theoremstyle{plain}

\theoremstyle{definition}

\theoremstyle{remark}

\usepackage[textsize=tiny]{todonotes}

\icmltitlerunning{Fly360: Omnidirectional Obstacle Avoidance within Drone View}

\begin{document}

\twocolumn[
  \icmltitle{Fly360: Omnidirectional Obstacle Avoidance within Drone View}



  \icmlsetsymbol{equal}{*}
  \begin{icmlauthorlist}
    \icmlauthor{Xiangkai Zhang}{equal,casia,insta}
    \icmlauthor{Dizhe Zhang}{equal,insta}
    \icmlauthor{Wenzhuo Cao}{insta}
    \icmlauthor{Zhaoliang Wan}{insta} \\
    \icmlauthor{Yingjie Niu}{insta}
    \icmlauthor{Lu Qi}{insta,whu}
    \icmlauthor{Xu Yang}{casia}
    \icmlauthor{Zhiyong Liu}{casia}
  \end{icmlauthorlist}
  \icmlaffiliation{casia}{Institute of Automation Chinese Academy of Sciences, Beijing, China}
  \icmlaffiliation{insta}{Insta360 Research, Shenzhen, China}
  \icmlaffiliation{whu}{Wuhan University,Wuhan, China}

  \icmlcorrespondingauthor{Lu Qi}{qqlu1992@gmail.com}
  \icmlcorrespondingauthor{Xu Yang}{xu.yang@ia.ac.cn}
  \icmlkeywords{Machine Learning, ICML}

  \vskip -0.5in
]



\printAffiliationsAndNotice{\icmlEqualContribution }
\begin{strip}
  \centering
  \vspace{-1pt}
  \includegraphics[width=\textwidth]{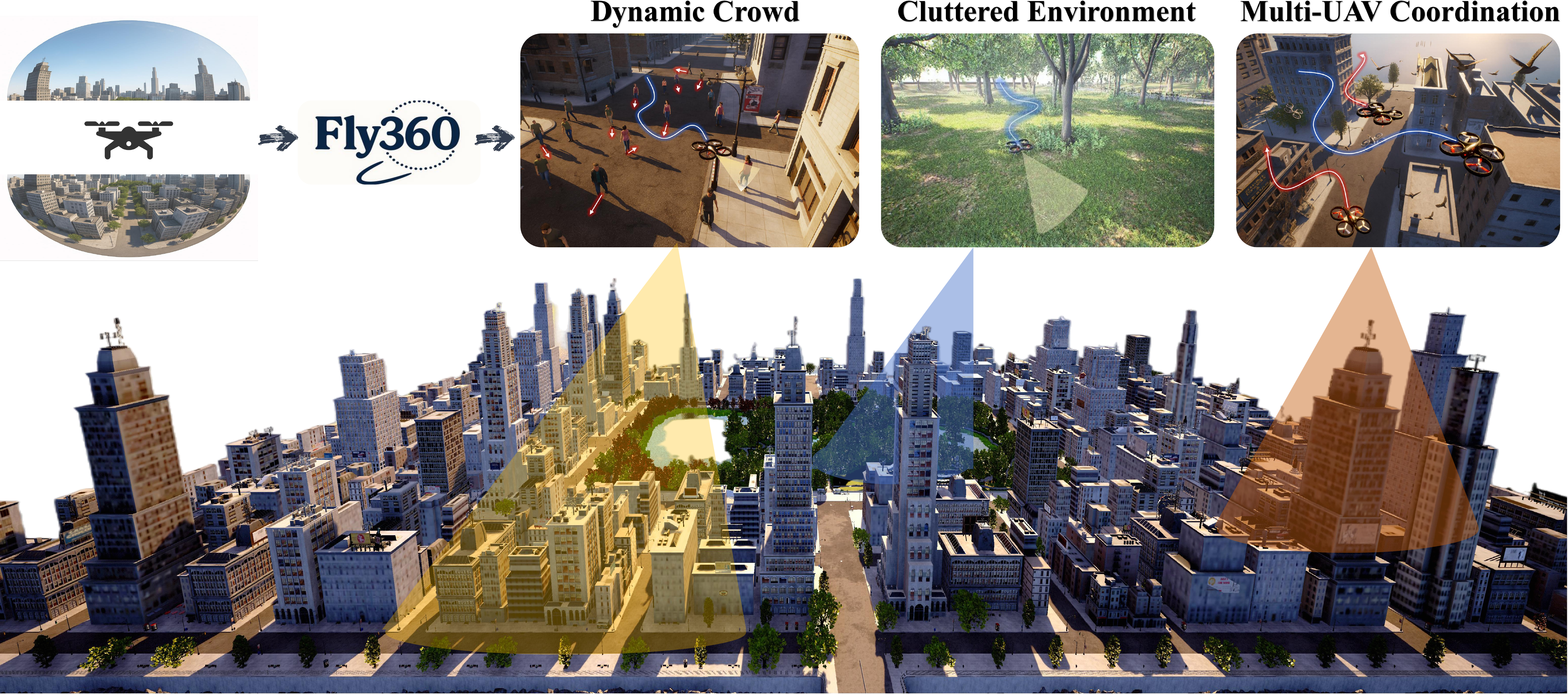}
  \captionof{figure}{We present \textbf{Fly360}, a panoramic-vision-based framework for omnidirectional UAV obstacle avoidance. 
  By mapping $360^{\circ}$ RGB inputs to control commands, Fly360 enables safe and agile navigation in complex environments without explicit mapping or specialized setups. 
  From dynamic crowds to cluttered natural scenes and multi-UAV coordination, our method can achieve omnidirectional awareness and robust flight beyond the limitations of forward-view sensing.}
  \label{fig:teaser}
  \vspace{-1pt}
\end{strip}

\begin{abstract}
Obstacle avoidance in unmanned aerial vehicles (UAVs), as a fundamental capability, has gained increasing attention with the growing focus on spatial intelligence.
However, current obstacle-avoidance methods mainly depend on limited field-of-view sensors and are ill-suited for UAV scenarios which require full-spatial awareness when the movement direction differs from the UAV’s heading.
This limitation motivates us
to explore omnidirectional obstacle avoidance for panoramic
drones with full-view perception.
We first study an underexplored problem setting in which a UAV must generate collision-free motion in environments with obstacles from arbitrary directions, and then construct a benchmark that consists of three representative flight tasks.
Based on such settings, we propose \textbf{Fly360}, a two-stage perception–decision pipeline with a fixed random-yaw training strategy.
At the perception stage, panoramic RGB observations are input and converted into depth maps as a robust intermediate representation.
For the policy network, it is lightweight and used to output body-frame velocity commands from depth inputs.
Extensive simulation and real-world experiments demonstrate that Fly360 achieves stable omnidirectional obstacle avoidance and outperforms forward-view baselines across all tasks.
Our model is available at \text{https://zxkai.github.io/fly360/}.
\vspace{-6pt}

\end{abstract}
\begin{figure*}[t]
    \centering
    \includegraphics[width=\textwidth]{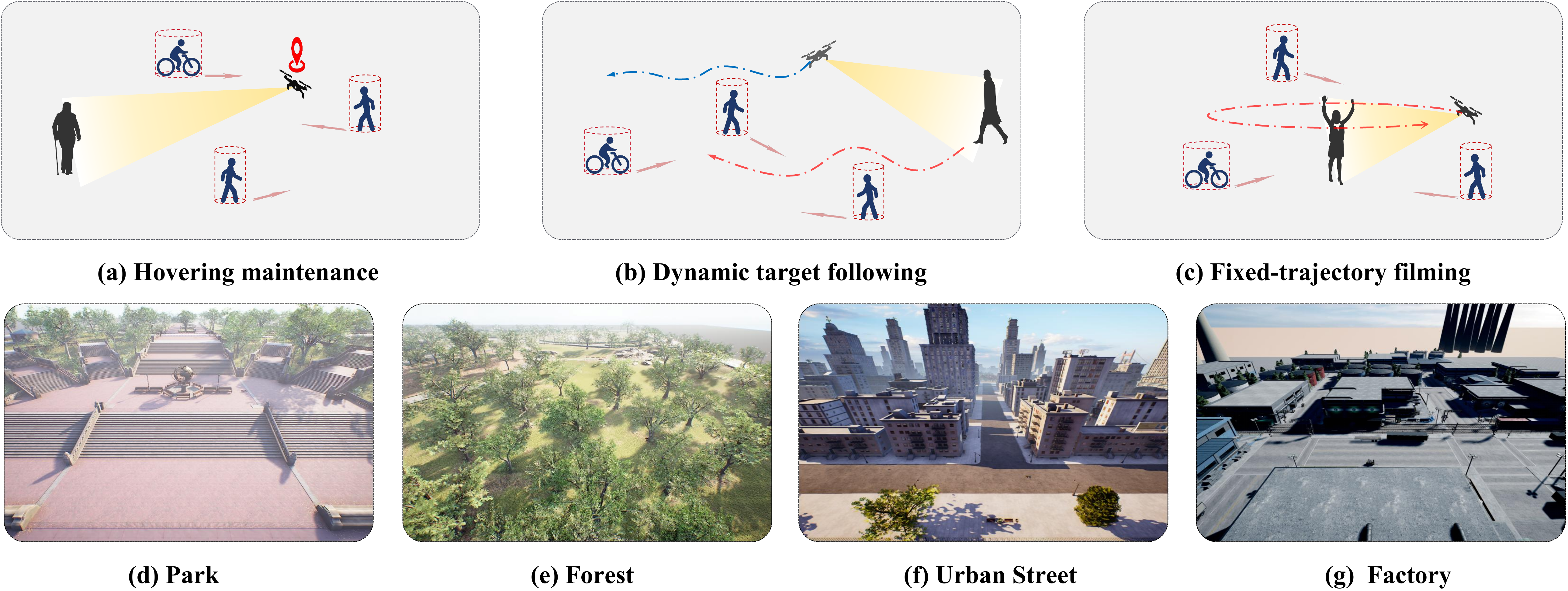}
    \caption{
    Overview of the experimental setting. 
    \textbf{Top:} The three representative tasks used to evaluate omnidirectional obstacle avoidance:(\textbf{a}) Hovering maintenance, where the UAV maintains a defined position and orientation while avoiding nearby obstacles; (\textbf{b}) Dynamic target following, where the UAV tracks a moving object while reacting to dynamic obstacles; and (\textbf{c}) Fixed-trajectory filming, where the UAV follows a predefined path around a target while maintaining camera orientation. 
    \textbf{Bottom:} The four high-fidelity simulation environments used in our evaluation, including (\textbf{d}) Park, (\textbf{e}) Forest, (\textbf{f}) Urban Street, and (\textbf{g}) Factory. 
    }
    \label{fig:task_env}
    \vspace{-10pt}
\end{figure*}
\section{Introduction}
\label{sec:intro}
Obstacle avoidance in unmanned aerial vehicles (UAVs)~\citep{cam,Liu23AerialVLN,ahmad2025future} has gained increasing attention with the growing focus on spatial intelligence. It serves as a basic task to support a wide range of applications, such as autonomous driving~\citep{autonomousdriving} and search-and-rescue~\citep{rescue,ge2025multi}.

Current obstacle-avoidance methods mainly depend on a limited field-of-view (FoV) captured by single or multiple cameras, using either traditional mapping–localization–planning–control pipelines~\citep{fastplanner,egoplanner,muticam1} or end-to-end learning-based approaches~\citep{wild,champion,newton}.
However, they are ill-suited for UAV scenarios, which require full-spatial awareness to support reliable motion and perception when the movement direction differs from the UAV’s heading.
Although Time-of-Flight sensors may be used, blind areas still remain, motivating us to explore omnidirectional obstacle avoidance for panoramic drones with full-view perception, where dual-fisheye setups have become practical, such as in Anti-Gravity products\cite{antigravity2025}.
To begin with, we identify an underexplored problem setting in which a UAV must generate collision-free motion in environments with dynamic obstacles from arbitrary directions. 
Therefore, the UAV’s motion should be decoupled from its heading, in contrast to previous limited-FoV settings where both of them are assumed to be aligned. 
By this motivation, we construct a benchmark that consists of three representative flight tasks, including hovering maintenance, dynamic target following, and fixed-trajectory filming.
As illustrated in Fig.~\ref{fig:task_env}, the UAV must maintain a desired orientation toward a main static or moving subject while hovering or moving along predefined or random trajectories.

Based on such settings, we propose Fly360, a two-stage perception–decision pipeline wrapped with a fixed random-yaw training strategy.

At the perception stage, we convert panoramic RGB images into depth maps using a pretrained panoramic depth model, which serves as a robust input for the policy network at the decision stage.
For the policy network, it is lightweight and used to
output body-frame velocity commands to control the motion of the UAV. 
Using depth as an intermediate representation helps alleviate the domain gap between training and validation, since most training pair-data is collected in simple simulations rather than the real world.
During policy training, the UAV is assigned a randomly sampled but fixed yaw angle for each episode in the simulator, encouraging the policy to learn orientation-invariant obstacle avoidance behaviors under arbitrary headings.

Last, the extensive experiments on the three proposed tasks demonstrate the effectiveness of our method.
For example, in the hovering maintenance task, Fly360 achieves success rates of up to 7/10 with cumulative collision times below 0.6 s, while all forward-view baselines fail in every setting with prolonged collisions exceeding 3–15 s. Consistent performance gains are also observed in the other two tasks, where Fly360 consistently attains higher success rates and lower collision times.
%
%
%
Our main contributions are summarized as follows:
\begin{itemize}

  \setlength{\itemsep}{0pt}
  \setlength{\parskip}{0pt}
  \setlength{\parsep}{0pt}
\item We formulate an underexplored omnidirectional obstacle-avoidance problem setting, where collision-free motion is generated under full-view perception with explicitly decoupled motion and heading.
\item We propose Fly360, a two-stage perception--decision framework with a fixed random-yaw training strategy, enabling orientation-invariant obstacle avoidance behaviors from panoramic observations.
\item We establish a benchmark with three representative task settings and validate the proposed method through extensive simulation and real-world experiments.
\end{itemize}
\section{Related Work}
\label{sec:relatedworks}

\subsection{UAV Obstacle-Avoidance Navigation}
Autonomous obstacle-avoidance flight has long been a core challenge in aerial robotics.
Early studies adopted a modular paradigm that separated the pipeline into perception~\cite{orb}, mapping~\cite{orbslam}, planning~\cite{fastplanner,egoplanner}, and control.
Such systems construct explicit maps from visual or range data, plan collision-free trajectories, and execute them through feedback controllers.
Representative works, including Fast-Planner~\cite{fastplanner} and EGO-Planner~\cite{egoplanner}, have achieved strong performance in dense environments.
However, modular designs suffer from cascading errors, inter-stage latency, and limited adaptability at high speeds or dynamic environments~\cite{review}.
These limitations have motivated a transition toward learning-based end-to-end frameworks that directly transfer sensory observations and UAV states to control outputs.

Early efforts such as CAD2RL~\cite{cad},
Fly by Crashing~\cite{crash}, and DroNet~\cite{dronet} established the feasibility of end-to-end learning, but their robustness and agility under complex or unseen conditions remained limited.
Significant progress followed with \cite{wild}, who demonstrated agile, high-speed flight in unknown cluttered environments.
\citet{champion} further advanced the field by achieving human-level performance in drone racing through deep reinforcement learning.
Recently,
\citet{newton} introduced differentiable rendering and physics to optimize policies directly from depth to action, while \citet{optical-flow} exploited optical flow as a compact motion representation for agile monocular flight.
\citet{transformeruav} explored Vision Transformers as unified perception encoders for UAV control.

Despite these advances, the perception of most end-to-end systems remains limited by the narrow FoV of forward-facing sensors.
As a result, these approaches struggle in scenarios that require omnidirectional obstacle avoidance.
Achieving a whole $360^{\circ}$ perception and navigation thus remains an open and essential topic for UAVs.
\subsection{Panoramic Visual Perception}
Panoramic vision enables comprehensive scene understanding by capturing omnidirectional visual information in a single observation.  ~\cite{panoreview,airsim360}
It offers a complete 360° field of view and removes blind spots, allowing agents to perceive cues from all directions.
Recent studies have explored panoramic perception in tasks such as semantic segmentation, depth estimation, and scene reconstruction~\cite{omnisam,onebev,omnidepth,unik3d,dap,dit360}.
Among these tasks, panoramic depth estimation has become a central topic for robotics, which can recover dense geometry from a single 360° image and provides essential depth cues for mapping and navigation.
Modern panoramic depth methods~\cite{daconv,daattention,DepthAnywhere} adapt network architectures to spherical geometry to handle projection distortion and maintain global consistency.  
Recent unified models such as UniK3D~\cite{unik3d} and MoGe~\cite{Moge} further generalize monocular geometry estimation across diverse camera types, extending applicability to wide-FoV and panoramic imagery.
Overall, current panoramic depth estimation approaches can achieve stable performance in scenes where extremely high precision is not required, forming a practical foundation for our work.

\begin{figure*}[!t]
  \centering
  \includegraphics[width=\textwidth]{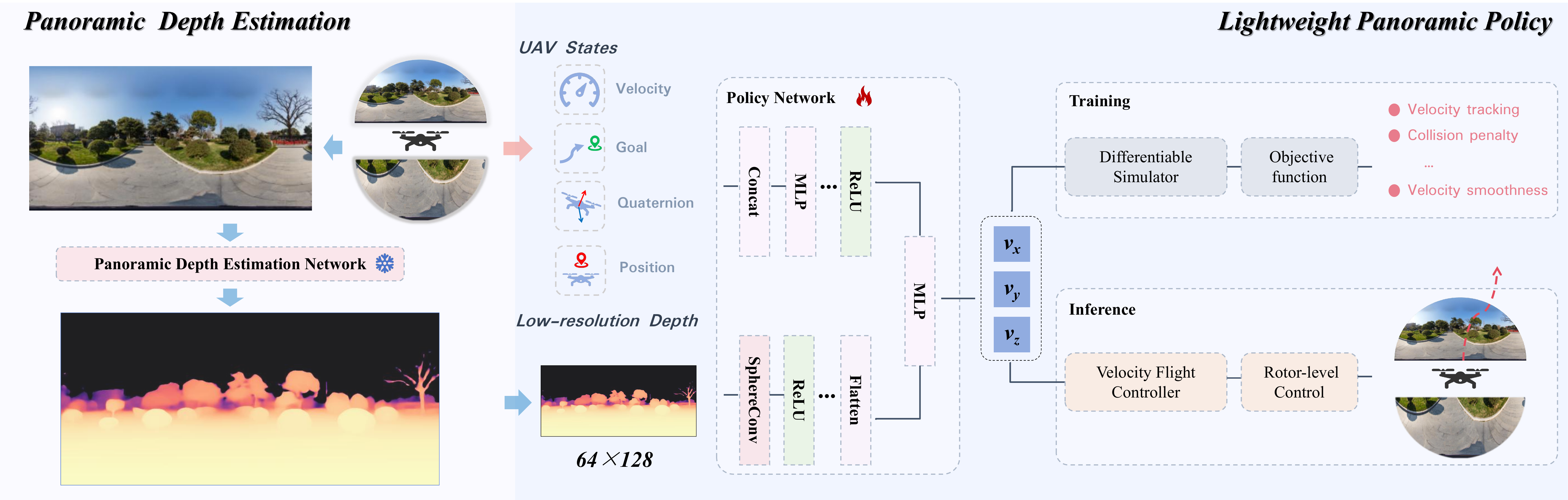}
  \caption{The proposed framework unifies panoramic perception and policy learning for omnidirectional UAV obstacle avoidance. Panoramic RGB observations are first processed by a panoramic depth estimation network to produce a depth map, which is then downsampled for policy inference. The policy network fuses low-resolution depth with UAV states to predict body-frame velocity commands. Training is conducted in a differentiable simulator while inference executes the predicted commands through the onboard velocity controller and rotor-level control. }
  \label{fig:pipeline}
  \vspace{-8pt}
\end{figure*}
\section{Methods}
The proposed \textbf{Fly360} takes $360^{\circ}$ panoramic RGB observations as input and outputs body-frame velocity commands that enable safe UAV flight.
In the following, we first formalize the considered problem, then describe the Fly360 system architecture, and finally present the training strategy used to obtain an orientation-invariant policy.
\subsection{Problem Formulation}
We address the task of panoramic vision based UAV navigation, where the vehicle is required to reach a given goal or follow a predefined trajectory while maintaining a desired orientation throughout the flight. 
This setting is representative of many real-world applications, such as aerial filming or inspection, where the UAV must move safely through cluttered environments while its heading remains fixed toward a target of interest.

At each time step $t$, the UAV captures a panoramic RGB image $I_t \in \mathbb{R}^{H \times W \times 3}$ in an equirectangular projection and obtains its current state
\begin{equation}
    \mathbf{s}_t = [\mathbf{p}_t, \mathbf{q}_t, \mathbf{v}_t],
\end{equation}
where $\mathbf{p}_t = [x_t, y_t, z_t]$ denotes the UAV position in the world frame, 
$\mathbf{q}_t$ represents its 3D orientation in quaternion form,
\begin{equation}
    \mathbf{q}_t = [w_t, x_t, y_t, z_t], \quad \text{with} \quad \|\mathbf{q}_t\| = 1,
\end{equation}
and $\mathbf{v}_t = [v_x, v_y, v_z]$ denotes the translational velocity in the body-frame. 
Given a goal position $\mathbf{g}$ or a sequence of waypoints $\{\mathbf{g}_i\}_{i=1}^N$, the objective is to generate a continuous control command defined as
\begin{equation}
    \mathbf{u}_t = \text{Fly360}(I_t, \mathbf{s}_t, \mathbf{g}),
\end{equation}
where $\mathbf{u}_t = [v_x, v_y, v_z]$ represents the desired velocity command in the body-frame. 
The $\text{Fly360}(\cdot)$ computes this command based on the current panoramic observation and state, 
driving the UAV safely toward the goal while avoiding obstacles perceived within the full $360^\circ$ field of view.

During execution, the velocity command $\mathbf{u}_t$ is combined with an external yaw control signal $\psi_c$, 
which may be manually specified or provided by a higher-level task module. 
Both $\mathbf{u}_t$ and $\psi_c$ are then transmitted to the low-level flight controller for actuation, 
where the controller fuses these inputs to produce the necessary rotor-level control actions.

\subsection{Fly360 System}

The proposed Fly360 system provides a two-stage framework that integrates panoramic perception and control policy learning for omnidirectional UAV obstacle avoidance under complex environmental and orientation constraints.
The system aims to map panoramic visual observations to body–frame velocity commands, enabling safe flight without relying on external sensors, explicit mapping, or handcrafted modules.
As illustrated in Fig.~\ref{fig:pipeline}, the Fly360 pipeline consists of a panoramic processing front-end and a lightweight control policy network.

\noindent\textbf{Panoramic perception and representation.}
To interpret panoramic inputs, the front-end converts each RGB panorama into a dense depth map using a pretrained panoramic depth model~\cite{unik3d}.
Rather than focusing on improving depth estimation itself, we emphasize its integration efficiency and robustness within the policy framework.
The depth is represented in a compact $64{\times}128$ equirectangular form and processed through spherical convolutions that preserve global geometric continuity while mitigating distortions near image boundaries.
This design provides omnidirectional geometry cues at low computational cost and serves as an effective intermediate representation. 

\begin{figure}[t]
  \centering
  \includegraphics[width=1\linewidth]{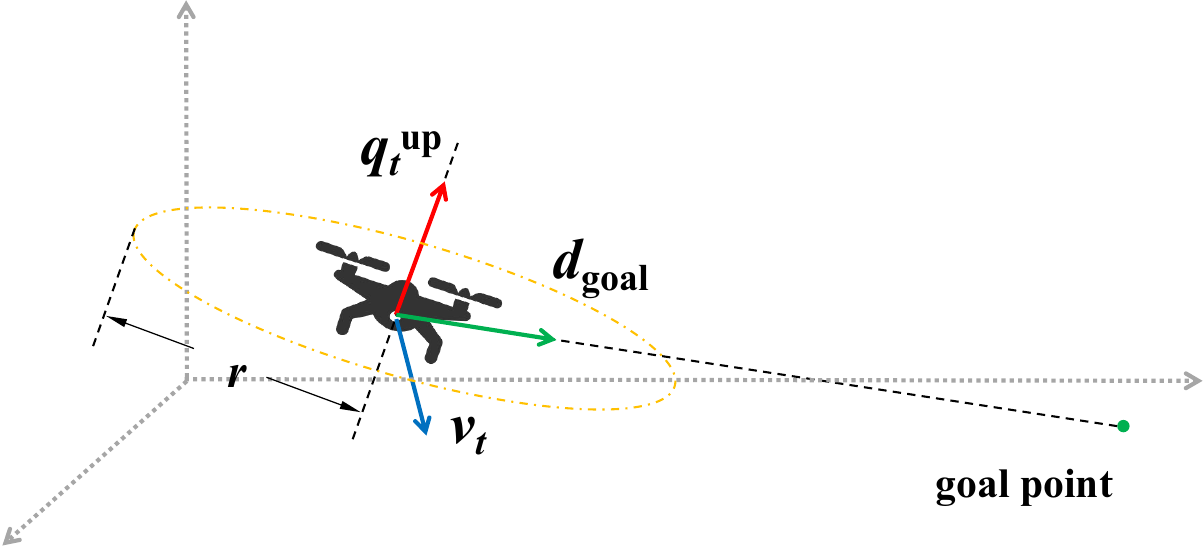}
  \caption{
  Illustration of the observation vector components used in the policy network. 
  The four components include the relative goal direction $\mathbf{d}_{\text{goal}}$, 
  current velocity $\mathbf{v}_t$, 
  upward orientation $\mathbf{q}^\text{up}_t$, 
  and predefined safety radius $r$.
  }
  \label{fig:observation_vector}
\end{figure}

\noindent\textbf{Panoramic policy.}
The control policy $\pi_\theta$ receives the panoramic depth map $D_t$ and an auxiliary observation vector $\mathbf{o}_t$, and predicts the corresponding body–frame velocity command:
\begin{equation}
\mathbf{u}_t = \pi_\theta(D_t, \mathbf{o}_t).
\end{equation}
The observation vector can be derived from the UAV state $\mathbf{s}_t$ and the goal position $\mathbf{g}$, including four components:
\begin{equation}
\mathbf{o}_t = \big[\, \mathbf{d}_{\text{goal}},\ \mathbf{v}_t,\ \mathbf{q}^\text{up}_t,\ r \,\big].
\end{equation}
Specifically, $\mathbf{d}_{\text{goal}} \in \mathbb{R}^3$ denotes the relative direction vector from the UAV to the next goal, 
$\mathbf{v}_t \in \mathbb{R}^3$ is the current body–frame velocity obtained from onboard state estimation, 
$\mathbf{q}^\text{up}_t \in \mathbb{R}^3$ represents the UAV’s upward orientation in the world frame to characterize its current attitude, 
and $r \in \mathbb{R}$ is a predefined safety radius of the UAV.
These components are concatenated and linearly projected to a 256-dimensional feature.
An illustration of the four components and their geometric relationships is provided in Fig.~\ref{fig:observation_vector}.

\noindent\textbf{Network architecture.}
The policy network employs a lightweight spherical convolutional recurrent design optimized for panoramic understanding and real-time control.
Two SphereConv~\cite{sphereconv} layers first extract globally consistent features from the equirectangular depth input, followed by several 2D convolutional blocks for hierarchical feature compression.
The encoded visual representation is concatenated with the projected observation vector embedding and passed through a single-layer GRU with 256 hidden units to model temporal dependencies in motion.
A linear head outputs the 3D velocity command $\mathbf{u}_t \in \mathbb{R}^3$.
This compact yet expressive architecture enables stable $360^{\circ}$ perception-to-control mapping and supports onboard deployment at real-time control frequencies.
Detailed layer configurations are provided in the Appendix.\ref{sec: Network Architecture Details}.
\subsection{Training Strategy}
\label{trainingstrategy}
Since the sim-to-real gap in depth is generally much smaller than that in RGB appearance, we use depth as the intermediate representation and train only the policy network from depth inputs, rather than jointly optimizing perception and control. This design reduces training difficulty, and it allows policy training to be conducted in a simple simulation environment without requiring high-fidelity visual realism. 
Moreover, the policy operates on aggressively downsampled panoramic depth inputs of only $64\times128$, further relaxing the requirement for depth map precision. 
The policy network is trained in a differentiable closed-loop simulator~\cite{newton}, which allows gradients to propagate through trajectory dynamics and avoids the instability and sample inefficiency of reinforcement learning. 

The policy is optimized using a combined objective that reflects the core requirements of panoramic obstacle avoidance navigation, which takes the form
\begin{equation}
\mathcal{L}
= \lambda_{\mathrm{trk}}\mathcal{L}_{\mathrm{trk}}
+ \lambda_{\mathrm{safe}}\mathcal{L}_{\mathrm{safe}}
+ \lambda_{\mathrm{smooth}}\mathcal{L}_{\mathrm{smooth}},
\label{eq:loss_overview}
\end{equation}
where the three components respectively promote navigation performance, obstacle-aware behavior, and dynamically feasible motion.  
All component definitions, weighting coefficients, and implementation details are provided in Appendix.\ref{sec:training details}.

\noindent\textbf{Orientation-invariant training.}
A key aspect of Fly360 is the proposed fixed random-yaw training strategy, designed to achieve orientation-invariant control. 
In forward-view obstacle avoidance, the training setup is straightforward. The UAV moves forward while its heading remains aligned with the motion direction, and obstacles are always encountered in the frontal region. Under this setting, a policy can be trained by repeatedly exposing it to variations of essentially the same scenario. As discussed in Sec.\ref{sec:intro}, this assumption no longer holds for omnidirectional obstacle avoidance. When motion direction and heading are decoupled, the UAV may encounter obstacles from arbitrary directions under different orientation constraints, making it infeasible to explicitly cover all possible training scenarios.

\begin{figure}[t]
  \centering
  \includegraphics[width=1\linewidth]{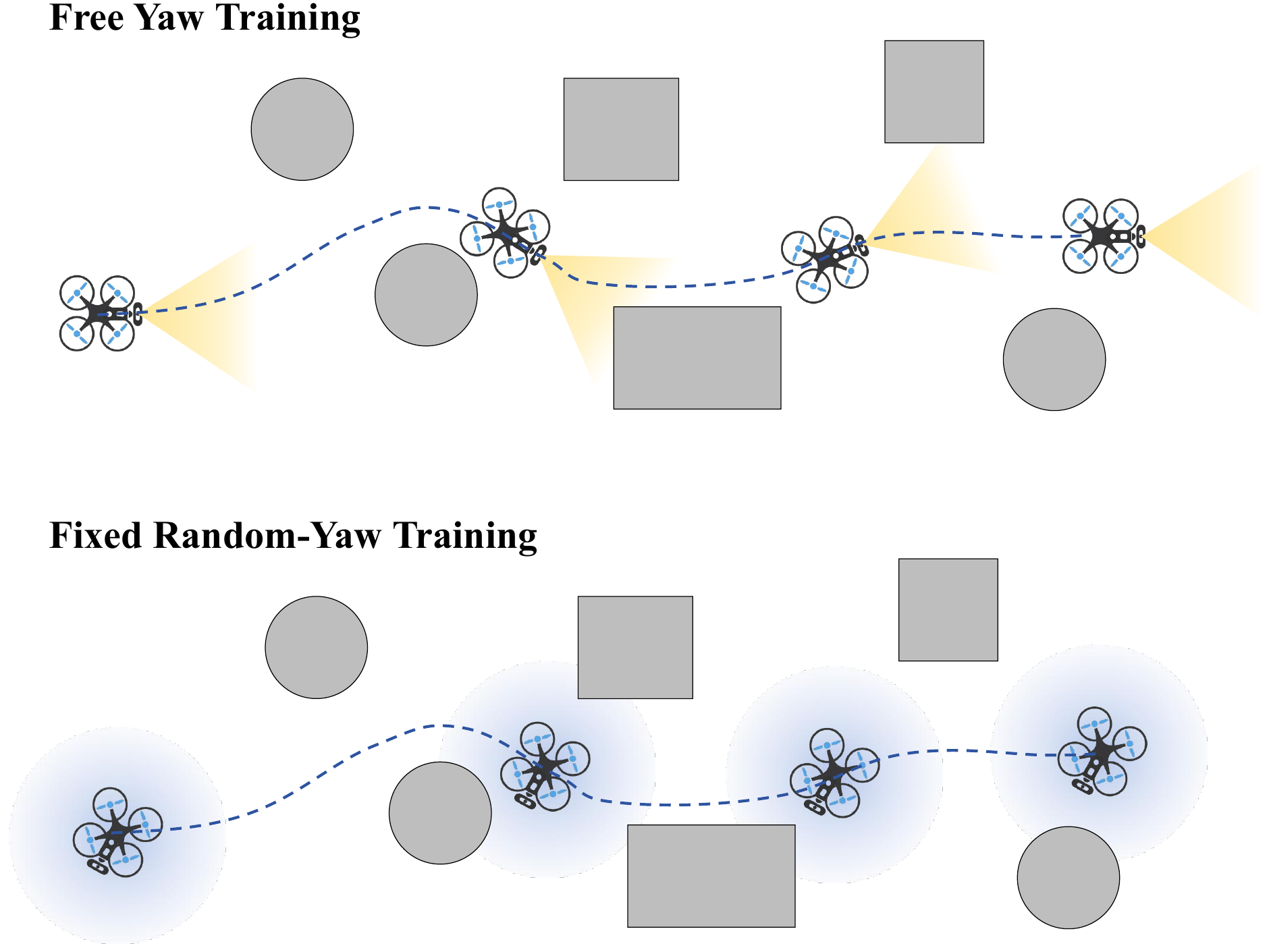}
  \caption{
  Illustration of the proposed \textbf{fixed random-yaw training} strategy. 
    In conventional \emph{free-yaw training} (top), the UAV’s yaw continuously aligns with the direction of motion, and the onboard camera (yellow cone) only observes a limited forward field of view. 
    In contrast, our training(bottom) randomly samples a yaw angle at the beginning of each rollout and keeps it constant throughout the episode. 
    The panoramic camera (blue region) provides a full $360^{\circ}$ field of view. 
  }
  \label{fig:training}
  \vspace{-10pt}
\end{figure}

Rather than enumerating these diverse scenarios, we focus on a more basic capability underlying omnidirectional obstacle avoidance. Regardless of the UAV’s heading, the desired obstacle avoidance behavior should remain consistent when responding to surrounding geometry. Once this capability is learned, the policy can generalize across scenarios without requiring exhaustive training coverage. Based on this observation, we propose a simple yet effective fixed random-yaw training strategy. By fixing a randomly sampled yaw angle throughout each episode, the policy is forced to 
interpret panoramic depth and establish a yaw-invariant mapping between omnidirectional geometry and collision-free motion, therefore enabling consistent obstacle avoidance behaviors independent of the UAV’s heading while still being trained under a simple and well-controlled simulation 
training setting.
Figure~\ref{fig:training} illustrates this training paradigm.

\begin{table*}[t]
\centering
\caption{
Quantitative results for \textbf{hovering maintenance} in \emph{park} and \emph{urban street} scenes.
Each entry reports success rate (SR) and collision time (CT, s) under two obstacle densities (3, 6) and two obstacle speeds (2.5, 5.0 m/s).
}
\label{tab:hover_results}

\tiny
\setlength{\tabcolsep}{2.5pt}         
\renewcommand{\arraystretch}{0.85}

\resizebox{0.9\textwidth}{!}{%
\begin{tabular}{llp{3.2cm}cccccccc}
\toprule
\multirow{4}{*}{Scene} 
& \multirow{4}{*}{View} 
& \multirow{4}{*}{Method} 
& \multicolumn{4}{c}{\#Objs = 3} 
& \multicolumn{4}{c}{\#Objs = 6} \\
\cmidrule(lr){4-7} \cmidrule(lr){8-11}
& & 
& \multicolumn{2}{c}{2.5 m/s} 
& \multicolumn{2}{c}{5.0 m/s}
& \multicolumn{2}{c}{2.5 m/s} 
& \multicolumn{2}{c}{5.0 m/s} \\
\cmidrule(lr){4-5} \cmidrule(lr){6-7} 
\cmidrule(lr){8-9} \cmidrule(lr){10-11}
& & 
& SR ($\uparrow$) & CT ($\downarrow$)
& SR ($\uparrow$) & CT ($\downarrow$)
& SR ($\uparrow$) & CT ($\downarrow$)
& SR ($\uparrow$) & CT ($\downarrow$) \\
\midrule

\multirow{7}{*}{\textbf{Park}}
& \multirow{2}{*}{Forward-view}
& ~\citet{newton}
& 0/10 & 3.48 & 0/10 & 3.45 & 0/10 & 5.18 & 0/10 & 5.11 \\
& 
& ~\citet{transformeruav} 
& 0/10 & 9.58 & 0/10 & 7.35 & 0/10 & 15.14 & 0/10 & 10.46 \\
\cmidrule(lr){2-11}
& \multirow{2}{*}{Multi-view}
& ~\citet{muticam1}
& 0/10 & 12.86 & 0/10 & 11.58 & 0/10 & 20.16 & 0/10 & 19.77 \\
& 
& ~\citet{muticam1}$^{\scriptscriptstyle *}$
& 0/10 & 1.11 & 0/10 & 1.33 & 0/10 & 1.49 & 0/10 & 2.45 \\
\cmidrule(lr){2-11}
& \multirow{2}{*}{Panoramic}

& \textbf{Ours} w/o fixed-yaw training 
& 3/10 & 1.11 & 1/10 & 1.60 & 0/10 & 3.18 & \textbf{3/10} & 4.85 \\
& 
& \textbf{Ours} 
& \textbf{6/10} & \textbf{0.13} & \textbf{7/10} & \textbf{0.54} 
& \textbf{1/10} & \textbf{0.90} & 1/10 & \textbf{1.84} \\

\midrule

\multirow{7}{*}{\textbf{Urban Street}}
& \multirow{2}{*}{Forward-view}
& ~\citet{newton}
& 0/10 & 4.80 & 0/10 & 6.40 & 0/10 & 5.30 & 0/10 & 7.60 \\
& 
& ~\citet{transformeruav} 
& 0/10 & 8.66 & 0/10 & 6.37 & 0/10 & 14.87 & 0/10 & 16.89 \\
\cmidrule(lr){2-11}
& \multirow{2}{*}{Multi-view}
& ~\citet{muticam1}
& 0/10 & 13.53 & 0/10 & 14.72 & 0/10 & 18.74 & 0/10 & 19.27 \\
& 
& ~\citet{muticam1}$^{\scriptscriptstyle *}$
& 0/10 & 1.00 & 0/10 & 1.44 & 0/10 & 2.63 & 0/10 & 1.97 \\
\cmidrule(lr){2-11}
& \multirow{2}{*}{Panoramic}
& \textbf{Ours} w/o fixed-yaw training 
& 3/10 & 1.19 & 3/10 & 3.35 & 0/10 & 4.41 & 0/10 & 4.28 \\
& 
& \textbf{Ours}  
& \textbf{7/10} & \textbf{0.09} & \textbf{3/10} & \textbf{1.27} 
& \textbf{4/10} & \textbf{0.62} & \textbf{2/10} & \textbf{1.56} \\

\bottomrule
\end{tabular}%
} 
\vspace{-4pt}
\end{table*}

\section{Experiments}

\subsection{Experimental Setup}

We evaluate the proposed Fly360 system through a series of experiments designed to assess its perception and obstacle-avoidance capabilities in both static and dynamic environments. The evaluation comprises three representative flight tasks that capture the core challenges of omnidirectional spatial geometry awareness and collision avoidance, as illustrated in Fig.~\ref{fig:task_env}(a)-(c).

\textbf{Task 1: Hovering Maintenance.} The UAV maintains a spatial position $(p_x, p_y, p_z)$ and yaw orientation $\psi_c$ toward a target. When obstacles approach, it must avoid them while maintaining a stable hover around the desired pose.

\textbf{Task 2: Dynamic Target Following.} The UAV tracks a moving target with a predefined relative offset (e.g., 5\,m in front) while keeping its yaw $\psi_c$ towards the target. To focus the evaluation on obstacle avoidance performance, the target position is provided as ground truth in this task.

\textbf{Task 3: Fixed-Trajectory Filming.} The UAV follows a given trajectory while keeping its camera oriented toward the target, avoiding obstacles that appear along its path.

These tasks together evaluate the system’s ability to achieve omnidirectional obstacle avoidance under different flight conditions. Our experiments are conducted in two stages. First, simulation tests are performed in a high-fidelity virtual environment to measure system performance under controlled conditions. Then, real-world flight tests are conducted on a physical UAV platform to verify the system’s transferability to practical environments.

\subsection{Simulation Experiments}
\label{sec:sim}

The simulation experiments quantitatively evaluate the robustness and effectiveness of \emph{Fly360} in diverse environmental conditions. All tests are carried out in the AirSim+UE4 simulator provided by AerialVLN~\cite{Liu23AerialVLN}, which reproduces realistic UAV dynamics, sensor feedback, and complex 3D obstacle configurations.

\textbf{Environments.}
Four representative environments are selected to capture different structural and visual characteristics: \emph{park}, \emph{forest}, \emph{urban street}, and \emph{factory}, as shown in Fig.~\ref{fig:task_env}(d)-(g). Each environment contains both static and dynamic obstacles. 

\noindent\textbf{Evaluation Protocol.}
Following~\cite{navrl,newton,transformeruav}, we adopt two quantitative metrics: 
Success Rate (SR), the ratio of trials completed without any collision; and 
Collision Time (CT) is the mean cumulative duration of collisions across all trials. 
Unlike conventional stop-on-impact settings, a trial continues after collision events, enabling the evaluation of post-collision recovery and whole trajectory stability. 
For $N$ trials, we compute
\begin{equation}
\mathrm{SR} = \frac{1}{N}\sum_{i=1}^{N} \mathbb{I}[\text{no collision in } i], \quad
\mathrm{CT} = \frac{1}{N}\sum_{i=1}^{N} c_i T^{\text{coll}}_i,
\end{equation}
where $T^{\text{coll}}_i$ is the total collision duration in trial $i$, and $c_i \in \{0,1\}$ indicates whether a collision occurred. All three tasks are evaluated over a fixed episode duration of 2 minutes, and a trial is considered successful only if the UAV completes the episode without any collisions.

\noindent\textbf{Baselines.}
We compare Fly360 with two types of baselines. The first group consists of state-of-the-art forward-view baselines proposed in~\citet{newton,transformeruav}, which both use a single front-facing depth map as input.
The second group includes multi-view baselines inspired by the multi-camera hardware setup in~\citet{muticam1}. Since the original work primarily focuses on hardware design and does not provide a learning-based control policy, we implement a learning-based method with a model structure similar to Fly360 based on its four fisheye-camera configuration. To ensure a fairer comparison in terms of perception coverage and learning difficulty induced by fisheye distortion, we further extend this setup to six camera views with $90^{\circ}$ FoV (front, back, left, right, up, and down) so that the multi-view baseline has perception comparable to that of panoramic input.
For fairness, all multi-view models are trained using the same fixed random-yaw training strategy as Fly360. All experiments are repeated ten times under randomized obstacle configurations to evaluate generalization and robustness. More model details on baselines are provided in Appendix.\ref{sec: Network Architecture Details}.

\noindent\textbf{Results and Discussion.}
Tables~\ref{tab:hover_results}–\ref{tab:track_results} jointly report the results across the three tasks and four environments. Overall, our method achieves the highest success rates and the lowest cumulative collision times in most settings, indicating that our framework is essential for stable omnidirectional obstacle avoidance in complex 3D scenes.

In the Hovering Maintenance task (Table~\ref{tab:hover_results}), the forward-view baselines completely fail, as they lack perception of obstacles approaching from the rear or lateral directions.
Consequently, these methods inevitably collide and remain trapped in cluttered environments, leading to prolonged collision times.
The multi-view setup improves over forward-view baselines by expanding perceptual coverage and providing partial awareness of surrounding obstacles, but the performance is unstable and varies across configurations. The four-camera fisheye setup still performs poorly as a forward-view baseline due to severe view distortion and fragmented perception, making it difficult to learn stable policies. The six-camera configuration achieves better results, but the overall performance remains limited.
A key reason is that depth is estimated independently for each view, which often leads to discontinuities and mismatches across views, especially when obstacles transition between camera frustums.
Moreover, different multi-view configurations require separate network architectures and dedicated training procedures, which inherently limit the generality and scalability of multi-view-based solutions.

Our panoramic policy achieves full spatial geometry awareness and effectively avoids collisions, as reflected by higher success rates and significantly shorter collision times.
The shorter collision times indicate that collisions are brief and infrequent, suggesting that the UAV can quickly recover to a stable hover rather than repeatedly contacting obstacles.
In contrast, very large CT values typically correspond to cases where the UAV becomes stuck against a static obstacle.

A similar trend is also observed in the Dynamic Target Following and Fixed-Trajectory Filming task as demonstrated in Table \ref{tab:follow_results} and \ref{tab:track_results}.
Forward and multi-view baselines often fail when obstacles appear laterally or from behind, resulting in prolonged contact and long collision times.
The panoramic method, in contrast, achieves notably lower CT and higher SR, demonstrating its ability to avoid obstacles omnidirectionally. 
Further experimental details and trajectory visualizations are included in the Appendix.\ref{sec: realworld exp}.

\begin{table}[t]
\centering
\caption{
Quantitative results for \textbf{dynamic target following} in \emph{forest} and \emph{factory} scenes.
Each entry reports the success rate (SR) and collision time (CT, s) under two target speeds (1.5, 3.0 m/s).
}
\label{tab:follow_results}

\tiny
\setlength{\tabcolsep}{2.2pt}
\renewcommand{\arraystretch}{1.0}

\resizebox{0.48\textwidth}{!}{%
\begin{tabular}{llp{2.4cm}cccc}
\toprule
\multirow{3}{*}{Scene}
& \multirow{3}{*}{View}
& \multirow{3}{*}{Method}
& \multicolumn{4}{c}{Target Speed} \\
\cmidrule(lr){4-7}
& & & \multicolumn{2}{c}{1.5 m/s} & \multicolumn{2}{c}{3.0 m/s} \\
\cmidrule(lr){4-5} \cmidrule(lr){6-7}
& & & SR($\uparrow$) & CT ($\downarrow$)& SR($\uparrow$) & CT($\downarrow$) \\
\midrule

\multirow{6}{*}{\textbf{Forest}}
& \multirow{2}{*}{Forward-view}
& ~\citet{newton}
& 1/10 & 10.19 & 0/10 & 6.39 \\
& & ~\citet{transformeruav}
& 0/10 & 38.90 & 0/10 & 60.20 \\
\cmidrule(lr){2-7}
& \multirow{2}{*}{Multi-view}
& ~\citet{muticam1}
& 0/10 & 29.60 &  0/10 & 27.45 \\
& & ~\citet{muticam1}$^{\scriptscriptstyle *}$
& 0/10 & 13.00 & 1/10 & 2.46 \\
\cmidrule(lr){2-7}
& \multirow{2}{*}{Panoramic}
& \textbf{Ours} w/o fixed-yaw training
& 0/10 & 2.19 & 2/10 & 1.10 \\
& & \textbf{Ours}
& \textbf{10/10} & \textbf{0} & \textbf{10/10} & \textbf{0} \\

\midrule

\multirow{6}{*}{\textbf{Factory}}
& \multirow{2}{*}{Forward-view}
& ~\citet{newton}
& 0/10 & 34.20 & 0/10 & 39.31 \\
& & ~\citet{transformeruav}
& 0/10 & 64.70 & 0/10 & 59.10\\
\cmidrule(lr){2-7}
& \multirow{2}{*}{Multi-view}
& ~\citet{muticam1}
& 0/10 & 27.85 & 0/10 & 33.45 \\
& & ~\citet{muticam1}$^{\scriptscriptstyle *}$
& \textbf{5/10} & 0.88 & 0/10 & 1.04 \\
\cmidrule(lr){2-7}
& \multirow{2}{*}{Panoramic}
& \textbf{Ours} w/o fixed-yaw training
& 0/10 & 57.73 & 0/10 & 33.40 \\
& & \textbf{Ours}
& \textbf{5/10} & \textbf{0.44} & \textbf{2/10} & \textbf{0.80} \\

\bottomrule
\end{tabular}
}
\end{table}

\begin{table}[t]
\centering
\caption{
Quantitative results for \textbf{fixed-trajectory filming} in \emph{park} and \emph{forest} scenes.
Each entry reports success rate (SR) and collision time (CT, s) under two obstacle speeds (3.0, 6.0~m/s).
}
\label{tab:track_results}

\tiny
\setlength{\tabcolsep}{2.2pt}
\renewcommand{\arraystretch}{1.0}

\resizebox{0.48\textwidth}{!}{%
\begin{tabular}{llp{2.4cm}cccc}
\toprule
\multirow{3}{*}{Scene}
& \multirow{3}{*}{View}
& \multirow{3}{*}{Method}
& \multicolumn{4}{c}{Obstacle Speed} \\
\cmidrule(lr){4-7}
& & & \multicolumn{2}{c}{3.0 m/s} & \multicolumn{2}{c}{6.0 m/s} \\
\cmidrule(lr){4-5} \cmidrule(lr){6-7}
& & & SR($\uparrow$) & CT ($\downarrow$)& SR($\uparrow$) & CT($\downarrow$) \\
\midrule

\multirow{6}{*}{\textbf{Park}}
& \multirow{2}{*}{Forward-view}
& ~\citet{newton}
& 1/10 & 52.04 & 0/10 & 54.02 \\
& & ~\citet{transformeruav}
& 0/10 & 45.89 & 0/10 & 41.68 \\
\cmidrule(lr){2-7}
& \multirow{2}{*}{Multi-view}
& ~\citet{muticam1}
& 0/10 & 57.19 & 0/10 & 73.70 \\
& & ~\citet{muticam1}$^{\scriptscriptstyle *}$
& 0/10 & 4.98 & 0/10 & 6.86 \\
\cmidrule(lr){2-7}
& \multirow{2}{*}{Panoramic}
& \textbf{Ours} w/o fixed-yaw training
& 0/10 & 59.48 & 0/10 & 35.33 \\
& & \textbf{Ours}
& \textbf{6/10} & \textbf{0.27} & \textbf{3/10} & \textbf{0.47} \\

\midrule

\multirow{6}{*}{\textbf{Forest}}
& \multirow{2}{*}{Forward-view}
& ~\citet{newton}
& 0/10 & 92.43 & 0/10 & 82.39 \\
& & ~\citet{transformeruav}
& 0/10 & 103.69 & 0/10 & 96.69 \\
\cmidrule(lr){2-7}
& \multirow{2}{*}{Multi-view}
& ~\citet{muticam1}
& 0/10 & 66.68 & 0/10 & 52.28 \\
& & ~\citet{muticam1}$^{\scriptscriptstyle *}$
& 0/10 & 2.74 & 0/10 & 3.79 \\
\cmidrule(lr){2-7}
& \multirow{2}{*}{Panoramic}
& \textbf{Ours} w/o fixed-yaw training
& 0/10 & 105.04 & 0/10 & 107.22 \\
& & \textbf{Ours}
& \textbf{10/10} & \textbf{0} & \textbf{10/10} & \textbf{0} \\

\bottomrule
\end{tabular}
}
\vspace{-6pt}
\end{table}
\noindent\textbf{Ablation Study and Robustness Analysis.}
The variant Ours w/o fixed-yaw training disables the proposed fixed random-yaw strategy during training. 
The performance degradation across all three tasks confirms that this strategy is critical for learning spatial geometry awareness and making orientation-invariant decisions that remain consistent under arbitrary headings. 
It enables the policy to establish a yaw-invariant mapping between omnidirectional spatial geometry and collision-free control. 

In addition, we conduct a robustness analysis to assess the 
policy’s sensitivity to inaccuracies in panoramic depth estimation. 
We add Gaussian noise to the predicted depth map to evaluate the policy’s sensitivity to depth estimation errors. Given a 
depth map $D$, we sample 
$\tilde{D} = D + \epsilon$, where 
$\epsilon \sim \mathcal{N}\!\big(0, (\gamma \,\bar{D})^2\big)$, 
$\bar{D}$ denotes the mean value of depth, and 
$\gamma \in \{0, 0.05,0.1, 0.2\}$ controls the relative noise level. 
As shown in Table \ref{tab:robustness}, Fly360 maintains stable avoidance performance across various noise levels, and even under the strongest perturbation, it continues to complete most trials with only a modest increase in collision time. 
More experiment results are provided in the Appendix.\ref{add_ablation}.
\begin{table}[t]
\centering
\caption{
Robustness of Fly360 to depth estimation errors under different noise levels.
Gaussian noise is added to the depth map.
}
\label{tab:robustness}
\resizebox{0.48\textwidth}{!}{%
\tiny
\setlength{\tabcolsep}{3.5pt}
\renewcommand{\arraystretch}{0.7}

\begin{tabular}{c cc cc cc}
\toprule
\multirow{2}{*}{$\sigma$} 
& \multicolumn{2}{c}{Filming (Forest)} 
& \multicolumn{2}{c}{Hovering (Park)} 
& \multicolumn{2}{c}{Following (Factory)} \\
\cmidrule(lr){2-3} \cmidrule(lr){4-5} \cmidrule(lr){6-7}
& SR $\uparrow$ & CT (s) $\downarrow$
& SR $\uparrow$ & CT (s) $\downarrow$
& SR $\uparrow$ & CT (s) $\downarrow$ \\
\midrule
0    & 10/10 & 0.00 & 1/10 & 0.90 & 10/10 & 0.00 \\
0.05 & 10/10 & 0.00 & 1/10 & 1.17 & 10/10 & 0.00 \\
0.10 & 9/10  & 0.07 & 2/10 & 1.01 & 10/10 & 0.00 \\
0.20 & 8/10  & 0.12 & 2/10 & 1.80 & 9/10 & 0.18 \\
\bottomrule
\end{tabular}
}
\vspace{-5pt}
\end{table}

\begin{table}[t]
\centering
\caption{
Comparison of model complexity and runtime performance on a desktop GPU (RTX~3090). 
The parameter count refers only to the policy network, while the latency and FPS reflect the total end-to-end system, including depth estimation and control.
}
\label{tab:runtime}
\footnotesize
\resizebox{0.48\textwidth}{!}{%
\setlength{\tabcolsep}{5pt}
\begin{tabular}{lcccc}
\toprule
Method & Params (M) & Latency (ms) & FPS (Hz) \\
\midrule
Forward-view~\cite{newton} & 2.1 & 21.0 & 47.6 \\
Forward-view~\cite{transformeruav} & 14.3 & 105.7 & 9.5 \\
Multi-view~\cite{muticam1} & 9.3 & 128.1 & 7.8 \\
Multi-view~\cite{muticam1}$^{\scriptscriptstyle *}$ & 9.2 & 130 & 7.7 \\
\textbf{Ours} (Panoramic) & 7.1 & 22.4 & 44.6 \\
\bottomrule
\end{tabular}
}
\vspace{-8pt}
\end{table}
\textbf{Runtime and Efficiency.}
Table~\ref{tab:runtime} summarizes the model complexity and inference speed of different methods, measured on a desktop GPU (RTX~3090). 
All models are tested under identical settings, with input resolutions of $448\times224$ for panoramic inputs and 6 or 4$\times224\times224$ for multi-view inputs.
The panoramic framework achieves comparable runtime to the forward-view baseline while providing significantly improved obstacle avoidance performance.

\begin{figure}[t]
  \centering
  \includegraphics[width=1\linewidth]{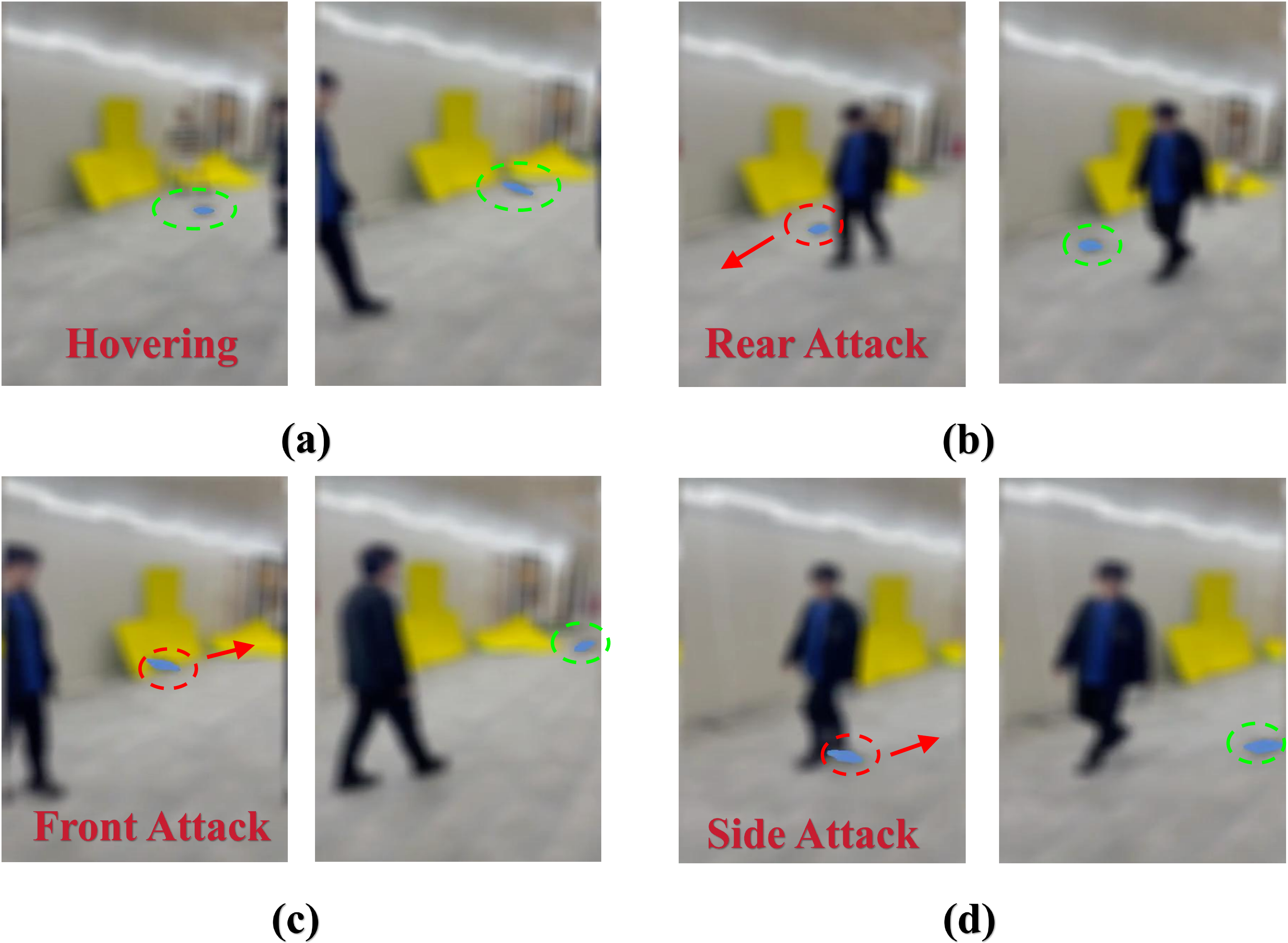}
  \caption{
  Real-world demonstration of omnidirectional avoidance during hovering.
  (a) Quadrotor platform equipped with two fisheye cameras whose images are automatically stitched into a panoramic view; due to commercial restrictions, detailed appearances are not shown.
  (b)-(d) Dynamic obstacle approaching from the rear, front, and side.
  }
  \label{realexp}
\vspace{-6pt}
\end{figure}
\subsection{Real-World Experiments}
\begin{figure}[!h]
  \centering
  \includegraphics[width=1\linewidth]{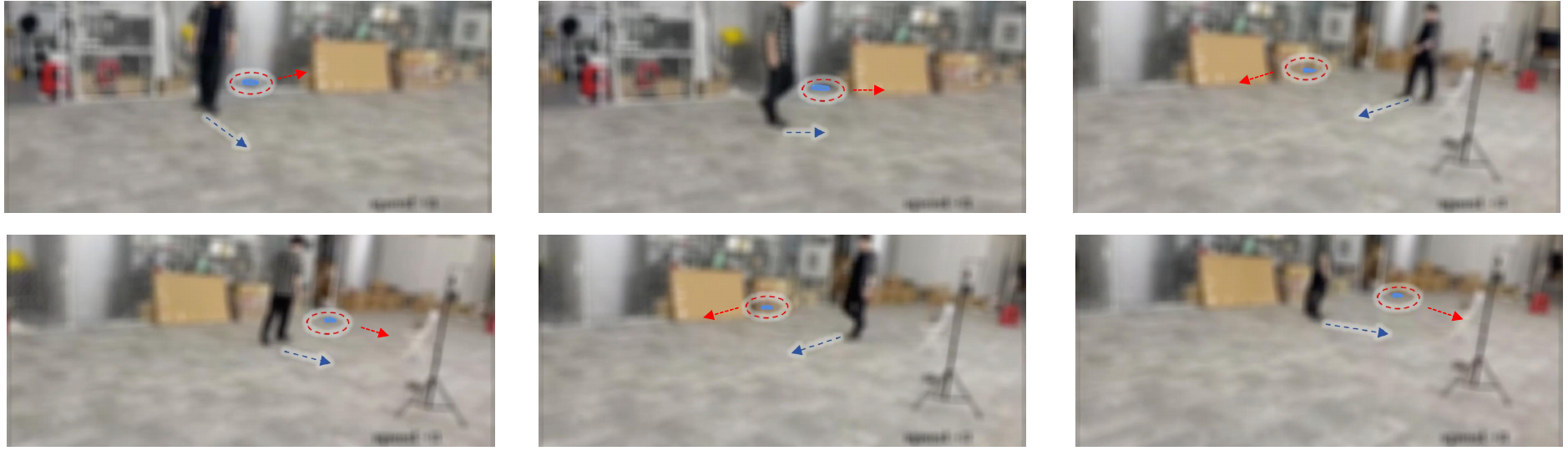}
  \caption{
    A challenging task scenario in which a human continuously pursues the UAV, demonstrating the UAV’s ability to evade a dynamic obstacle consistently. 
  }
  \label{chasing}
\end{figure}

To validate the proposed framework in practical conditions, we deploy the Fly360 system on a custom quadrotor platform equipped with a panoramic sensor. 
Onboard attitude sensing provides the vehicle state, and the policy outputs body–frame velocity commands that are transmitted to the flight controller.
As shown in Fig.~\ref{realexp}, in a confined hovering scenario with dynamic obstacles approaching, Fly360 consistently achieves omnidirectional avoidance and recovers stable hovering, demonstrating reliable sim-to-real transfer.
Furthermore, in a more challenging chasing experiment (Fig.~\ref{chasing}), where a human continuously pursues the UAV, the system enables sustained collision-free flight under persistent and unpredictable dynamic threats, highlighting the robustness of the proposed framework in real-world environments.
Quantitative results and latency analysis are reported in Table.\ref{tab:realworld_latency}.
Additional results are provided in the Appendix.\ref{sec: realworld exp} and supplementary video.
\section{Conclusion}
This work has presented \textbf{Fly360}, a two-stage framework for omnidirectional UAV obstacle avoidance based on panoramic vision perception.
By integrating panoramic depth estimation and policy network learning within a single framework, the system maps panoramic visual observations to control commands, enabling responsive control in complex environments.
The proposed yaw-fixed training strategy further strengthens spatial geometry awareness, allowing the policy to maintain consistent obstacle-avoiding performance under arbitrary flight headings.
Overall, Fly360 offers a practical solution for vision-based omnidirectional UAV navigation, advancing robust and efficient autonomous flight in dynamic real-world environments.
In future work, we will further explore improvements in real-time efficiency and generalization across environments.

\begin{table}[!t]
\centering
\caption{
Quantitative results and latency of Fly360 in real-world experiments.
P: perception (RGB$\rightarrow$depth), D: decision (policy inference), C: control interface.
}
\label{tab:realworld_latency}
\resizebox{0.48\textwidth}{!}{
\renewcommand{\arraystretch}{0.82}
\footnotesize
\begin{tabular}{cccc cccc}
\toprule
\multicolumn{4}{c}{Simple Hovering} 
& \multicolumn{4}{c}{Challenging Chasing} \\
\cmidrule(lr){1-4} \cmidrule(lr){5-8}
SR $\uparrow$  & P (ms) & D (ms) & C (ms) 
& SR $\uparrow$ & P (ms) & D (ms) & C (ms)  \\
\midrule
5/5 & 60 & 12 & 18
& 3/5 & 57 & 10 & 21  \\
\bottomrule
\end{tabular}}
\vspace{-10pt}
\end{table}

\section*{Impact Statement}
This paper presents work whose goal is to advance the field of machine learning for autonomous aerial systems, specifically UAV omnidirectional obstacle avoidance under panoramic perception. 
There are potential societal consequences associated with autonomous systems, but we do not believe this work raises ethical concerns beyond those commonly studied in the field.


\bibliography{example_paper}
\bibliographystyle{icml2026}

\clearpage
\newpage
\appendix

\section*{Appendix Overview}

This supplementary document provides additional materials that complement the main paper and support the clarity and reproducibility of our work. The contents are organized into the following components:
\begin{itemize}
    \item \textbf{Network Architecture Details.}  
    We provide complete architectural specifications for our panoramic policy network. Detailed configurations of baseline models used for comparison are also included.
    \item \textbf{Training Details.}  
    This section presents extended training information, including complete loss formulations, coefficient definitions, optimization hyperparameters, rollout settings, and other implementation details that were omitted from the main text for brevity.
    \item \textbf{Additional Experimental Results.}  
    We report extended simulation outcomes, visualization, and robustness analyses that further validate the effectiveness of Fly360. Additional real-world qualitative results are also provided to illustrate the system’s performance under diverse physical conditions.
    \item \textbf{Supplementary Video.}  
    A supplementary video is also provided, showcasing representative trajectory visualizations, avoidance behaviors, and real-world experiments to better illustrate the capabilities and practical performance of Fly360.
\end{itemize}
\section{Network Architecture Details}
\label{sec: Network Architecture Details}
The Fly360 adopts a lightweight spherical–convolutional recurrent architecture that is specifically designed for panoramic perception and real-time UAV control. This section provides the complete architectural specifications of the Fly360 policy network as well as baseline policies used for comparison. All table formats follow a unified structure to clearly illustrate the differences among the three type of models.
\subsection{Fly360 Panoramic Policy Network}
The Fly360 policy network receives a $64{\times}128$ equirectangular depth map together with an auxiliary observation vector 
$\mathbf{o}_t = [\mathbf{d}_{\text{goal}}, \mathbf{v}_t, \mathbf{q}^{\text{up}}_t, r]$.
The panoramic depth input is first processed by two SphereConv layers that extract globally consistent $360^{\circ}$ geometric features. A series of 2D convolutional layers further compresses the representation into a compact visual embedding. Both the visual embedding and the projected observation embedding are mapped into 256-dimensional vectors and fused before being fed into a single-layer GRUCell. The GRU maintains short-term temporal memory and produces the hidden state used by the final linear layer to predict the 3D body–frame velocity command $\mathbf{u}_t = [v_x, v_y, v_z]$.

The full architecture is summarized in Table~\ref{tab:policy_arch}.
\begin{table}[t]
  \centering
  \footnotesize
  \caption{
  Architecture of the Fly360 panoramic policy network. 
  The model receives a $64{\times}128$ equirectangular depth map and outputs a 3D body–frame velocity command.
  }
  \vspace{2pt}
  \resizebox{0.48\textwidth}{!}{%
  \begin{tabular}{lccc}
    \toprule
    \textbf{Layer} & \textbf{Operation} & \textbf{Config} & \textbf{Output Size} \\
    \midrule
    0  & Input                    & ERP depth (1 ch)                 & $1 \times H \times W$ \\
    1  & SphereConv2d + LeakyReLU & (32, $3{\times}3$, s=2)           & $32 \times H_1 \times W_1$ \\
    2  & SphereConv2d + LeakyReLU & (64, $3{\times}3$, s=2)           & $64 \times H_2 \times W_2$ \\
    3  & Conv2d + LeakyReLU       & (64, $3{\times}3$, s=1)           & $64 \times H_3 \times W_3$ \\
    4  & Conv2d + LeakyReLU       & (64, $2{\times}2$, s=2)           & $64 \times H_4 \times W_4$ \\
    5  & Conv2d + LeakyReLU       & (128, $3{\times}3$, s=1)          & $128 \times H_5 \times W_5$ \\
    6  & Conv2d + LeakyReLU       & (128, $3{\times}3$, s=1)          & $128 \times H_6 \times W_6$ \\
    7  & Flatten                  & --                                & $D_{\text{flat}}$ \\
    8  & Linear (visual proj)     & $(D_{\text{flat}} \rightarrow 256)$ & $256$ \\
    9  & Linear (obs proj)        & $(9 \rightarrow 256)$               & $256$ \\
    10 & GRUCell                  & hidden = 256                      & $256$ \\
    11 & Linear (control head)    & $(256 \rightarrow 3)$              & $3$ \\
    \bottomrule
  \end{tabular}}
  \label{tab:policy_arch}
\end{table}
This design provides a compact yet expressive panoramic policy network suitable for onboard deployment. The network maintains strong omnidirectional geometric understanding while keeping computation affordable for real-time control.

\subsection{Forward-View Baseline}

The forward-view baseline~\cite{newton} follows a monocular depth–based control architecture that is widely adopted in prior UAV navigation research. Its design is based on convolutional–recurrent policies widely used in vision-based flight, and therefore serves as a strong single-camera baseline for comparison. The model processes a single forward-facing depth image with a $90^{\circ}$ field of view using a lightweight ConvNet encoder. The extracted visual embedding is combined with a projected observation embedding and propagated through a GRUCell. The final linear head predicts the body–frame velocity command.
The complete architecture is listed in Table~\ref{tab:forward_arch}.

\begin{table}[h]
  \centering
  \footnotesize
  \caption{
  Architecture of the forward-view baseline network.
  The model receives a single-view depth image and outputs the 3D velocity command.
  }
  \vspace{2pt}
  \resizebox{0.48\textwidth}{!}{%
  \begin{tabular}{lccc}
    \toprule
    \textbf{Layer} & \textbf{Operation} & \textbf{Config} & \textbf{Output Size} \\
    \midrule
    0  & Input                  & Depth (1 ch)                   & $1{\times}H{\times}W$ \\
    1  & Conv2d + LeakyReLU     & (32, $2{\times}2$, s=2)        & $32{\times}H_1{\times}W_1$ \\
    2  & Conv2d + LeakyReLU     & (64, $3{\times}3$, s=1)        & $64{\times}H_2{\times}W_2$ \\
    3  & Conv2d + LeakyReLU     & (128, $3{\times}3$, s=1)       & $128{\times}H_3{\times}W_3$ \\
    4  & Flatten                & --                             & $D_{\text{flat}}$ \\
    5  & Linear (visual proj)   & $(D_{\text{flat}} \rightarrow 192)$ & $192$ \\
    6  & Linear (obs proj)      & $(9 \rightarrow 192)$           & $192$ \\
    7  & GRUCell                & hidden = 192                   & $192$ \\
    8  & Linear (control head)  & $(192 \rightarrow 3)$          & $3$ \\
    \bottomrule
  \end{tabular}}
  \label{tab:forward_arch}
\end{table}
The second forward-view baseline~\cite{transformeruav} adopts a lightweight vision-transformer (ViT) encoder to replace the convolutional visual frontend, which improves representation capacity while retaining an efficient control head. Following the official implementation, the input depth image is first resized to $60{\times}90$, then processed by a two-stage MixTransformer encoder to extract multi-scale visual features. The features are spatially aligned and fused through upsampling and pixel-shuffle operations, followed by a linear projection to a compact latent vector. This latent is concatenated with the low-dimensional proprioceptive observation (e.g., velocity) and a quaternion state, and finally mapped to the 3D body-frame velocity command using either an LSTM controller (\emph{ViT+LSTM}) or a lightweight MLP head (\emph{ViT+FC}). The complete architecture is summarized in Table~\ref{tab:vit_arch}.
\begin{table}[t]
\centering
\footnotesize
\setlength{\tabcolsep}{4pt}
\renewcommand{\arraystretch}{1.05}
\caption{
Architecture of the transformer-based forward-view baseline~\cite{transformeruav}.
The model receives a single-view depth image and outputs the 3D velocity command.
}
\label{tab:vit_arch}
\vspace{2pt}
\resizebox{0.48\textwidth}{!}{%
\begin{tabular}{lccc}
\toprule
Layer & Operation & Config & Output Size \\
\midrule
0  & Input + Resize
   & Depth (1 ch) $\rightarrow 60{\times}90$
   & $1{\times}60{\times}90$ \\
1  & MixTransformer Stage-1
   & $(1\rightarrow 32)$, patch=7, stride=4, pad=3, heads=1, layers=2
   & $32{\times}H_1{\times}W_1$ \\
2  & MixTransformer Stage-2
   & $(32\rightarrow 64)$, patch=3, stride=2, pad=1, heads=2, layers=2
   & $64{\times}H_2{\times}W_2$ \\
3  & Feature Align
   & PixelShuffle($\times 2$) on Stage-2 + Upsample Stage-1 $\rightarrow 16{\times}24$
   & $48{\times}16{\times}24$ \\
4  & Conv2d (downsample)
   & $(48\rightarrow 12)$, $3{\times}3$, pad=1
   & $12{\times}16{\times}24$ \\
5  & Flatten + Linear (visual proj)
   & $(4608 \rightarrow 512)$
   & $512$ \\
6  & Concat (state)
   & $[512,\; \mathbf{o}/10,\; \mathbf{q}]$
   & $517$ \\
7a & LSTM Controller (ViT+LSTM)
   & 3 layers, hidden=128, dropout=0.1
   & $128$ \\
8a & Linear (control head)
   & $(128 \rightarrow 3)$
   & $3$ \\
\midrule
7b & MLP Head (ViT+FC)
   & Linear$(517\rightarrow 256)$ + LeakyReLU
   & $256$ \\
8b & Linear (control head)
   & $(256 \rightarrow 3)$
   & $3$ \\
\bottomrule
\end{tabular}%
}
\vspace{-10pt}
\end{table}

\subsection{Multi-View Baseline}
The multi-view baseline extends the forward-view setting by providing a policy processing six synchronized perspective depth maps corresponding to the front, back, left, right, up, and down directions or four fisheye-type depth maps similar to ~\citet{muticam1}. These multi-directional views are stacked along the channel dimension and processed by a deeper ConvNet encoder that increases representational capacity. The resulting feature map is flattened and projected to a 384-dimensional latent vector. The observation vector is projected to the same dimensionality, and the fused embedding is passed into a GRUCell before the final linear prediction head. This baseline serves as a strong multi-camera policy.
The whole architecture is summarised in Table~\ref{tab:multiview_arch}.

\begin{table}[h]
  \centering
  \footnotesize
  \caption{
  Architecture of the multi-view baseline network. 
  The model receives six perspective depth maps and outputs the 3D body–frame velocity command.
  }
  \vspace{2pt}
  \resizebox{0.48\textwidth}{!}{%
  \begin{tabular}{lccc}
    \toprule
    \textbf{Layer} & \textbf{Operation} & \textbf{Config} & \textbf{Output Size} \\
    \midrule
    0  & Input                  & Depth (6 or 4 ch)                   & $C{\times}H{\times}W$ \\
    1  & Conv2d + LeakyReLU     & (256, $3{\times}3$, s=2)       & $256{\times}H_1{\times}W_1$ \\
    2  & Conv2d + LeakyReLU     & (128, $3{\times}3$, s=1)       & $128{\times}H_2{\times}W_2$ \\
    3  & Conv2d + LeakyReLU     & (128, $3{\times}3$, s=2)       & $128{\times}H_3{\times}W_3$ \\
    4  & Conv2d + LeakyReLU     & (128, $3{\times}3$, s=1)       & $128{\times}H_4{\times}W_4$ \\
    5  & Conv2d + LeakyReLU     & (128, $3{\times}3$, s=2)       & $128{\times}H_5{\times}W_5$ \\
    6  & Conv2d + LeakyReLU     & (256, $3{\times}3$, s=1)       & $256{\times}H_6{\times}W_6$ \\
    7  & Flatten                & --                             & $D_{\text{flat}}$ \\
    8  & Linear (visual proj)   & $(D_{\text{flat}} \rightarrow 384)$ & $384$ \\
    9  & Linear (obs proj)      & $(9 \rightarrow 384)$          & $384$ \\
    10 & GRUCell                & hidden = 384                   & $384$ \\
    11 & Linear (control head)  & $(384 \rightarrow 3)$          & $3$ \\
    \bottomrule
  \end{tabular}}
  \label{tab:multiview_arch}
\end{table}
\noindent\textbf{Summary.} 
All policies adopt a similar structure composed of a visual encoder, an observation projection module, a recurrent unit, and a linear control head for predicting body–frame velocity commands.
The forward-view baselines process a single forward-facing depth image using either convolutional or transformer-based encoders, providing perception within a limited field of view.
The multi-view baseline increases perceptual coverage by jointly processing multiple synchronized depth maps from different directions, at the cost of handling view-wise representations.
Fly360 instead operates on a unified $360^{\circ}$ equirectangular depth map and applies SphereConv-based feature extraction to obtain an omnidirectional visual representation within the same recurrent control framework.
\section{Training Details}
\label{sec:training details}
This section provides additional implementation details for the policy optimization process that complement the training strategy described in the main paper. As stated in the main text, only the policy network is trained. The panoramic depth estimator remains frozen because of domain gap between training and validation, since most training data is collected in simple simulations. The focus of training is therefore on learning robust panoramic control rather than depth estimation.
\subsection{Training Environment and Control Loop}
Training is performed in a differentiable closed-loop simulator based on~\cite{newton}. The simulator models the UAV as a point-mass system with thrust-based actuation and first-order attitude dynamics. At the beginning of each training iteration, the environment is reset, and a rollout of $T=600$ control steps is executed.
A key aspect of the simulator is that the control interval is not fixed. Instead, the effective control timestep $\Delta t$ is sampled at every step:
\begin{equation}
\Delta t \sim \mathcal{N}\left({1}/{15},\; {0.1}/{15}\right),
\end{equation}
which produces a time-varying control frequency around $15\,\mathrm{Hz}$. This stochastic timing variation approximates the frequency jitter observed on real UAV platforms and improves the robustness of the learned controller to non-uniform execution rates. The simulator integrates the dynamics using the sampled timestep and returns updated depth observations and states to the policy.
The GRU hidden state is reset at the beginning of every rollout, and all models (Fly360 and the two baselines) share identical simulation settings, including stochastic control timing, batch size, rollout length, and environment parameters.

At each timestep, the policy receives the preprocessed depth map together with an observation vector extracted from the current vehicle state. The model predicts a body-frame velocity command, which is then transformed into the world frame through the current rotation matrix. This command contributes to the thrust update through the environment dynamics. Temporal consistency is maintained by a recurrent hidden state that is updated at every step.
\subsection{Training and Optimization Setup}

All policies, including Fly360 and baselines, are trained on a single NVIDIA RTX~3090 GPU (24\, GB). Training is performed using the AdamW optimizer with an initial learning rate of $10^{-3}$, followed by a cosine annealing schedule that decays the learning rate to $1\%$ of its initial value. Gradients are updated after each rollout by averaging the loss across all $600$ simulation steps. To ensure fair comparison, all models share identical simulator configurations, rollout settings, and optimization hyperparameters, and all coefficients in the objective function match the values specified in the training script. Each policy is trained until convergence, which typically requires $5$--$10$ thousand gradient steps, corresponding to approximately $2$--$6$ hours of training time on the RTX~3090 GPU.
\subsection{Detailed Objective and Hyperparameters}
In the main paper, the learning objective is presented in a compact form,
\begin{equation}
\mathcal{L}
= \lambda_{\mathrm{trk}}\mathcal{L}_{\mathrm{trk}}
+ \lambda_{\mathrm{safe}}\mathcal{L}_{\mathrm{safe}}
+ \lambda_{\mathrm{smooth}}\mathcal{L}_{\mathrm{smooth}},
\end{equation}
which summarizes the three high-level components required for panoramic obstacle-avoidance navigation: 
(1)~velocity tracking, 
(2)~safety around obstacles, 
and (3)~smooth, dynamically feasible control.

For completeness, we provide here the complete loss formulation used in our implementation.  
These terms are expanded versions of the three components above; the additional notation does not contradict the main paper but decomposes each part into finer-grained penalties used in the training script.

\noindent\textbf{Velocity tracking.}
This corresponds to $\mathcal{L}_{\mathrm{trk}}$ in the main paper.  
We track the smoothed executed velocity $\overline{\mathbf{v}}_t$ toward the goal-directed target velocity $\mathbf{v}^\star_t$:
\begin{equation}
\mathcal{L}_{\mathrm{trk}}
= \mathrm{SmoothL1}\!\left(
\overline{\mathbf{v}}_t - \mathbf{v}_t^{\star}
\right),
\end{equation}
where  
$\overline{\mathbf{v}}_t = \frac{1}{30}\sum_{k=0}^{29}\mathbf{v}_{t-k}$ is a 30-step moving average of the executed velocity,  
and $\mathbf{v}^\star_t$ is the goal-directed target velocity after magnitude saturation.

\noindent\textbf{Safety.}
The safety term $\mathcal{L}_{\mathrm{safe}}$ is implemented using two complementary penalties:
\begin{align}
\mathcal{L}_{\mathrm{avoid}}
&= (1-d_t)_{+}^{2}\, v_{\mathrm{to\_pt}},\\
\mathcal{L}_{\mathrm{collide}}
&= \mathrm{softplus}(-\gamma d_t)\, v_{\mathrm{to\_pt}},
\end{align}
where  
$d_t$ is the clearance to the nearest obstacle after subtracting a safety margin,  
$v_{\mathrm{to\_pt}}$ is the approach velocity toward the obstacle,  
and $\gamma$ controls the steepness of the collision barrier.  
These two terms respectively penalize reduced clearance and entering the unsafe region.

\noindent\textbf{Smoothness.}
The high-level smoothness term $\mathcal{L}_{\mathrm{smooth}}$ is expanded into regularization on acceleration and jerk:
\begin{equation}
\mathcal{L}_{\mathrm{acc}}=\|\mathbf{a}_t\|_2^2,\qquad
\mathcal{L}_{\mathrm{jerk}}=\|\mathbf{j}_t\|_2^2,
\end{equation}
where  
$\mathbf{a}_t$ and $\mathbf{j}_t$ denote acceleration and jerk computed from successive control inputs.  

\noindent\textbf{Auxiliary consistency.}
An additional self-supervised objective improves training stability:
\begin{equation}
\mathcal{L}_{\mathrm{vpred}} = \|\hat{\mathbf{v}}_t - \mathbf{v}_t\|_2^2,
\end{equation}
where $\hat{\mathbf{v}}_t$ is the predicted instantaneous velocity and $\mathbf{v}_t$ is the executed one.  
Other auxiliary terms (e.g., directional bias, ground affinity) appear in the training script but have zero weight and are therefore omitted.

\noindent\textbf{Final Optimization Objective}
Combining all active components yields the full training objective:
\begin{equation}
\begin{aligned}
\mathcal{L} = \;&
\lambda_{\mathrm{trk}}\, \mathcal{L}_{\mathrm{trk}}
+ \lambda_{\mathrm{safe}}
  \left( \mathcal{L}_{\mathrm{avoid}} + \mathcal{L}_{\mathrm{collide}} \right)
\\&+
\lambda_{\mathrm{smooth}}
  \left( \mathcal{L}_{\mathrm{acc}} + \mathcal{L}_{\mathrm{jerk}} \right)
+
\lambda_{\mathrm{vp}}\, \mathcal{L}_{\mathrm{vpred}}.
\end{aligned}
\end{equation}
Table~\ref{tab:supp_hyperparam} lists all the exact coefficients used in training.  
\begin{table}[t]
\centering
\caption{Loss coefficients and related hyperparameters used in the training script.}
\label{tab:supp_hyperparam}
\renewcommand{\arraystretch}{1.2}  
\begin{tabular}{lccc}
\toprule
Category & Term & Coefficient / Parameter & Value \\
\midrule
Tracking 
& $\mathcal{L}_{\mathrm{trk}}$ 
& $\lambda_{\mathrm{trk}}$ 
& $1.0$ 
\\[-1pt]
\hdashline
\noalign{\vskip 3pt}  

Safety 
& $\mathcal{L}_{\mathrm{avoid}}$
& $\lambda_{\mathrm{safe}}$(avoid)
& $1.5$ 
\\
& $\mathcal{L}_{\mathrm{collide}}$ 
& $\lambda_{\mathrm{safe}}$(collide) 
& $2.0$ 
\\
& --- 
& $\gamma$
& $32$ 
\\[-1pt]
\hdashline
\noalign{\vskip 3pt}

Smoothness 
& $\mathcal{L}_{\mathrm{acc}}$ 
& $\lambda_{\mathrm{smooth}}$ (acc)
& $0.01$ 
\\
& $\mathcal{L}_{\mathrm{jerk}}$ 
& $\lambda_{\mathrm{smooth}}$ (jerk)
& $0.001$ 
\\[-1pt]
\hdashline
\noalign{\vskip 3pt}

Auxiliary 
& $\mathcal{L}_{\mathrm{vpred}}$ 
& $\lambda_{\mathrm{vp}}$
& $2.0$ 
\\
\bottomrule
\end{tabular}
\end{table}
\begin{figure*}[t]
  \centering
  \includegraphics[width=\textwidth]{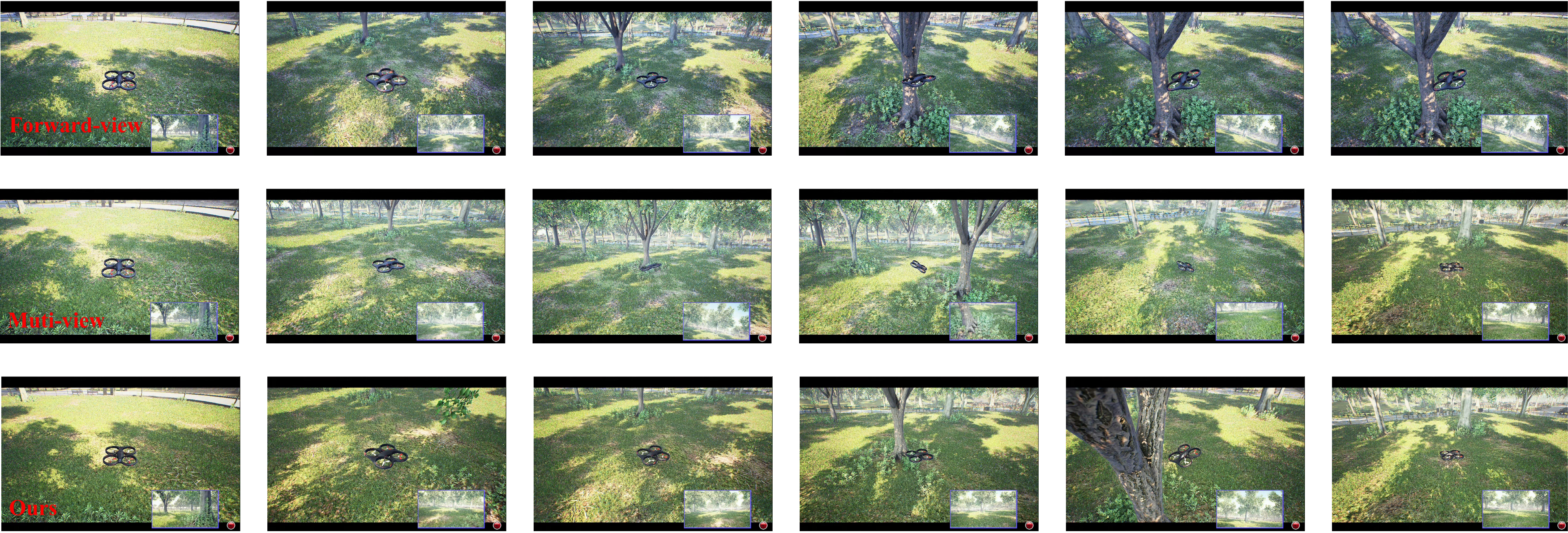}
  \caption{
  \textbf{Fixed-trajectory filming in a dense forest.}
  Representative frames from forward-view, multi-view, and Fly360 policies. 
  Fly360 follows the predefined filming path while maintaining collision-free motion, whereas the baselines exhibit frequent lateral failures or inconsistent obstacle responses.
  }
  \label{fig:forest_traj}
\end{figure*}
\begin{figure*}[t]
  \centering
  \includegraphics[width=\textwidth]{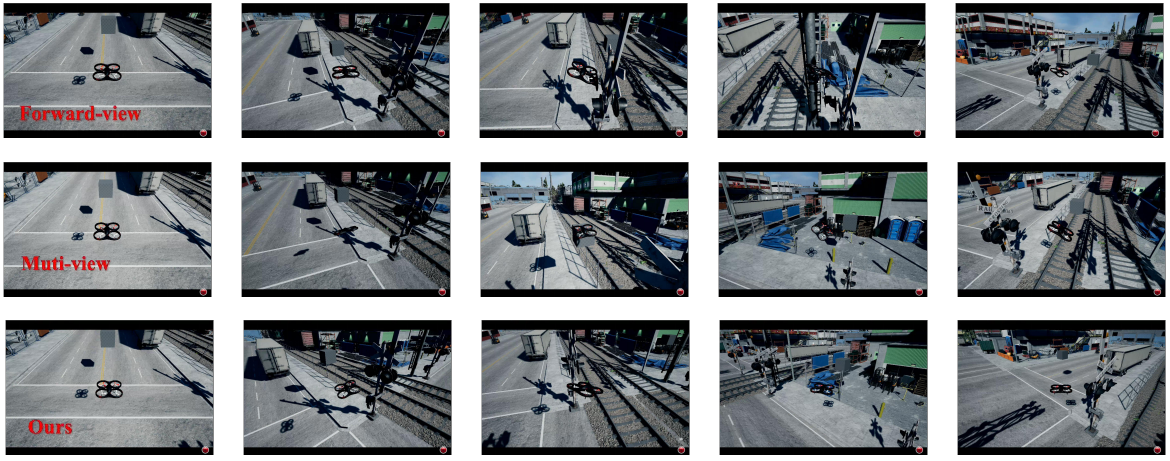}
  \caption{
  \textbf{Dynamic target following in an industrial environment.}
  Fly360 maintains stable tracking and consistent obstacle clearance, while forward-view and multi-view baselines frequently collide in tight spaces.
  }
  \label{fig:factory_follow}
\end{figure*}
\begin{figure*}[t]
  \centering
  \includegraphics[width=\textwidth]{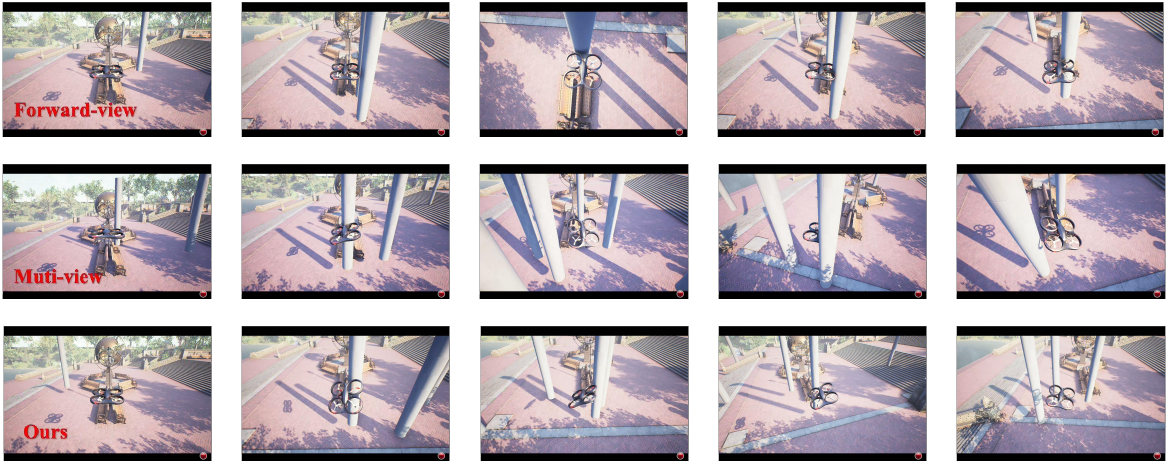}
  \caption{
  \textbf{Hovering maintenance in a park environment.}
  Fly360 reacts promptly to obstacles approaching from any direction and consistently returns to the target hover point, while the baselines show delayed or incomplete avoidance responses.
  }
  \label{fig:park_hover}
\end{figure*}
\begin{table*}[t]
\centering
\caption{
Quantitative results for \textbf{hovering maintenance} in \emph{park} and \emph{urban street} scenes.
Each entry reports success rate (SR) and collision time (CT, s) under two obstacle densities (3, 6) and two obstacle speeds (2.5, 5.0 m/s).
}
\label{tab:hover_results_ab}

\tiny
\setlength{\tabcolsep}{2.5pt}
\renewcommand{\arraystretch}{0.9}

\resizebox{0.9\textwidth}{!}{%
\begin{tabular}{lp{3.6cm}cccccccc}
\toprule
\multirow{4}{*}{Scene} 
& \multirow{4}{*}{Method} 
& \multicolumn{4}{c}{\#Objs = 3} 
& \multicolumn{4}{c}{\#Objs = 6} \\
\cmidrule(lr){3-6} \cmidrule(lr){7-10}
& & 
\multicolumn{2}{c}{2.5 m/s} 
& \multicolumn{2}{c}{5.0 m/s}
& \multicolumn{2}{c}{2.5 m/s} 
& \multicolumn{2}{c}{5.0 m/s} \\
\cmidrule(lr){3-4} \cmidrule(lr){5-6} 
\cmidrule(lr){7-8} \cmidrule(lr){9-10}
& & 
SR $\uparrow$ & CT $\downarrow$
& SR $\uparrow$ & CT $\downarrow$
& SR $\uparrow$ & CT $\downarrow$
& SR $\uparrow$ & CT $\downarrow$ \\
\midrule

\multirow{3}{*}{\textbf{Park}}
& Ours (joint depth--policy training)
& 0/10 & 39.20 & 0/10 & 41.33 & 0/10 & 44.17 & 0/10 & 44.28 \\
& Ours w/o fixed-yaw training
& 3/10 & 1.11 & 1/10 & 1.60 & 0/10 & 3.18 & \textbf{3/10} & 4.85 \\
& \textbf{Ours}
& \textbf{6/10} & \textbf{0.13} & \textbf{7/10} & \textbf{0.54}
& \textbf{1/10} & \textbf{0.90} & 1/10 & \textbf{1.84} \\

\midrule

\multirow{3}{*}{\textbf{Urban Street}}
& Ours (joint depth--policy training)
& 0/10 & 26.30 & 0/10 & 31.40 & 0/10 & 20.13 & 0/10 & 32.17 \\
& Ours w/o fixed-yaw training
& 3/10 & 1.19 & 3/10 & 3.35 & 0/10 & 4.41 & 0/10 & 4.28 \\
& \textbf{Ours}
& \textbf{7/10} & \textbf{0.09} & \textbf{3/10} & \textbf{1.27}
& \textbf{4/10} & \textbf{0.62} & \textbf{2/10} & \textbf{1.56} \\

\bottomrule
\end{tabular}}
\vspace{-6pt}
\end{table*}

\section{Additional Experimental Results}
\label{sec: realworld exp}
This section provides supplementary results and extended evidence that complement the simulation and real-world evaluations presented in the main paper. We expand upon the trajectory visualizations, ablation results, and robustness analysis in simulation, and further provide additional qualitative results from real-world experiments.

\subsection{Simulation Trajectory Visualizations}
This part provides extended qualitative results that complement the simulation evaluations presented in the main paper. We present trajectory visualizations for three representative tasks-—Fixed-Trajectory Filming, Dynamic Target Following, and Hovering Maintenance—-together with additional comparisons against the forward-view(~\citet{newton}) and multi-view(~\citet{muticam1}$^{\scriptscriptstyle *}$) baselines. These visualizations illustrate how panoramic perception improves stability, obstacle awareness, and robustness across diverse environments. Complete trajectories and dynamic behaviors are further demonstrated in the supplementary video.

\noindent\textbf{Fixed-Trajectory Filming in Forest Environment}
Figure~\ref{fig:forest_traj} compares Fly360 with the two baselines in a dense forest environment where the UAV is required to follow a predefined camera path. The forward-view method frequently loses situational awareness in lateral and rear directions and often collides when trees enter the blind zone. The multi-view method improves coverage but still suffers from inconsistent perception across view boundaries, which usually leads to more divergent and large-magnitude avoidance maneuvers that lack stability. In contrast, Fly360 maintains smooth progress along the trajectory and reliably avoids surrounding trees using a unified panoramic representation.

\noindent\textbf{Dynamic Target Following in Industrial Environment}
Figure~\ref{fig:factory_follow} shows qualitative results for dynamic target following in a cluttered industrial scene. This task emphasizes responsiveness to a moving target while avoiding static obstacles such as machinery, beams, and pillars. The forward-view baseline often becomes trapped when the target exits the narrow frontal field of view. The multi-view policy responds faster but still struggles in sudden occlusion or tight passages. Fly360 consistently keeps the target in view and maintains safe separation from obstacles, demonstrating stable tracking behavior even under target motion.

\noindent\textbf{Hovering Maintenance in Park Environment}
Figure~\ref{fig:park_hover} visualizes the performance of the three policies in a hovering maintenance task within a semi-open park environment. While holding a position relative to a nearby wall and surrounding vegetation, obstacles approach from multiple directions. The forward-view controller detects only frontal hazards, and the multi-view policy exhibits delayed reactions to side and rear intrusions. Fly360 demonstrates omnidirectional awareness, performing short evasive maneuvers before returning smoothly to the hover point.

\noindent\textbf{Summary.}  
Across all three tasks, the visualizations illustrate that panoramic perception enables the UAV to maintain stable trajectories, avoid collisions proactively, and recover effectively from local disturbances. Complete sequences, including full-length trajectories and dynamic avoidance behaviors, are provided in the supplementary video.
\begin{table}[t]
\centering
\caption{
Robustness of Fly360 under varying obstacle sizes in the hovering maintenance task.
All experiments are conducted in the same environment with obstacle speed fixed at 5.0~m/s.
Each entry reports the average collision time (CT, s) over ten trials.
}
\label{tab:obstacle_size}

\setlength{\tabcolsep}{6pt}
\renewcommand{\arraystretch}{1.0}

\resizebox{0.48\textwidth}{!}{%
\begin{tabular}{cccccccc}
\toprule
Obstacle Size $r$ (m)
& 0.01 & 0.05 & 0.10 & 0.20 & 0.30 & 0.40 & 0.50 \\
\midrule
CT (s) $\downarrow$
& 4.10 & 3.25 & 2.83 & 2.20 & 3.43 & 2.23 & 2.88 \\
\bottomrule
\end{tabular}}
\vspace{-10pt                            }
\end{table}

\subsection{Additional Ablation Analysis}
\label{add_ablation}
To examine the necessity of the proposed two-stage perception–decision design, we conduct a framework with joint training of the depth estimator and policy network.
Due to the limited visual diversity and simplified structures in the simulation training, the jointly trained model fails to converge and performs poorly, even underperforming forward-view baselines as shown in Table.\ref{tab:hover_results_ab}.

In addition, to assess whether the aggressive depth downsampling adopted in our framework leads to performance degradation, we conduct an auxiliary experiment by varying the obstacle size while keeping other settings unchanged. This experiment directly evaluates the policy’s sensitivity to fine-grained geometric details. The quantitative results, summarized in Table~\ref{tab:obstacle_size}, show that the proposed method maintains stable obstacle avoidance performance across different obstacle scales, indicating that depth downsampling does not introduce significant performance loss.

\subsection{Additional Real-World Results}
We provide more qualitative examples from real-world indoor experiments on our project website to further validate the proposed system. In numerous trials, the system consistently avoids collisions and returns to its target hover position with minimal drift, demonstrating effective sim-to-real transfer.

Additionally, we test a challenging scenario in which a human continuously pursues the UAV, and the detailed visualization is provided in the supplementary video. This test demonstrates the UAV's ability to consistently evade a dynamic human target, further highlighting the robustness and adaptability of the system in complex, real-world environments. The system remains stable and responsive even in the presence of partial occlusions, fast-approaching obstacles, and visually ambiguous backgrounds, supporting the real-world feasibility and reliability of the proposed panoramic navigation framework.


\end{document}